\documentclass{article}

\usepackage[preprint]{neurips_2026}

\makeatletter
\renewcommand{\@notice}{}
\makeatother

\usepackage[utf8]{inputenc}
\usepackage[T1]{fontenc}
\usepackage{hyperref}
\hypersetup{hypertexnames=false,hidelinks}
\usepackage{url}
\usepackage{booktabs}
\usepackage{tabularx}
\usepackage{array}
\usepackage{amsmath}
\usepackage{amssymb}
\usepackage{amsfonts}
\usepackage{nicefrac}
\usepackage{microtype}
\usepackage{xcolor}
\usepackage{graphicx}
\usepackage{tikz}
\usepackage{algorithm}
\usepackage{algpseudocode}

\usetikzlibrary{arrows.meta,positioning,calc}

\title{
Towards Reversible Forgetting:
Managing Obsolete Knowledge in Continual Enterprise AI Agents
}

\author{
Nilutpaul Sarker Yash
\And
Tirtho Roy
\And
Ushashi Bhattacharjee
}

\begin{document}
\raggedbottom

\maketitle


\begin{abstract}
Continual learning has traditionally treated forgetting as a failure, emphasizing preservation of previously acquired knowledge as environments evolve. We argue that this objective is incomplete for enterprise AI agents operating in non-stationary environments, where customers, policies, tools, workflows, regulations, and market conditions change over time. Indiscriminate retention can allow obsolete knowledge to influence decisions, creating negative transfer and operational risk. We therefore propose \textbf{reversible forgetting}: a conceptual framework with three operational memory states---active, dormant, and retired---and a reactivation transition that can restore dormant knowledge when its relevance returns. We instantiate the framework as a \textbf{Hysteretic Reversible Memory Controller} that accumulates relevance evidence, uses asymmetric thresholds to prevent state oscillation, tests reactivation in shadow mode, and gates retirement through policy. The framework reduces the influence of obsolete information without conflating temporary suppression with permanent erasure. Finance illustrates the idea: knowledge useful under one market regime may become harmful under another yet regain relevance when similar conditions recur.
\end{abstract}


\noindent\textbf{TL;DR:} Enterprise agents should reversibly suppress obsolete knowledge rather than retain or erase it indiscriminately; HRMC operationalizes this position through auditable memory states, hysteresis, shadow-tested reactivation, and policy-gated retirement.


\section{Introduction}

Continual learning asks how a system can acquire knowledge without losing useful prior capabilities~\citep{parisi2019continual,delange2022survey,wang2024comprehensive}. For enterprise agents, however, older knowledge should not always remain operational: customers, policies, tools, workflows, regulations, and markets change.

This creates a second failure mode beyond catastrophic forgetting: obsolete workflows, strategies, policies, or experiences can cause negative transfer when retrieval continues to expose them. Lifelong-agent research identifies memory evolution as an open problem~\citep{zheng2025roadmap}, and insight-governance work shows that stale verbal experience can harm later decisions~\citep{cui2026closing}.

\paragraph{Position.}
Continual enterprise agents should support \textbf{reversible forgetting}: obsolete or contextually irrelevant knowledge should be suppressible from normal use without necessarily being erased, while preserving provenance and later reactivation.

This reframes continual learning as management of a knowledge lifecycle rather than maximization of retention. Our argument makes four contributions: (1) we distinguish catastrophic forgetting from intentional, beneficial suppression; (2) we introduce active, dormant, and retired memory states connected by auditable suppression and reactivation transitions; (3) we propose a concrete controller with temporal evidence accumulation, hysteresis, shadow reactivation, and policy-gated retirement; and (4) we outline a research and benchmarking agenda for detection, granularity, retention, reactivation, evaluation, safety, and governance.


\section{From catastrophic forgetting to beneficial forgetting}

Continual learning studies sequential adaptation while maintaining useful earlier capabilities~\citep{wickramasinghe2024review,ven2024catastrophic}. Its central failure is \textbf{catastrophic forgetting}: new learning damages knowledge that remains valuable. Enterprise agents face the converse risk. A redesigned system, deprecated tool, superseded rule, or changed operating regime can make preserved knowledge misleading. Thus, accidental loss of useful knowledge is catastrophic, but controlled suppression of obsolete knowledge can be adaptive. The question becomes \textbf{what should influence the agent now, remain dormant, or be retired}.

\subsection{Relationship to prior and adjacent work}

Catastrophic-forgetting methods preserve useful capabilities~\citep{kirkpatrick2017overcoming}, while backward transfer improves old tasks using new knowledge~\citep{lin2022backward}; our concern is whether old knowledge should influence the current task at all. Machine unlearning removes designated training influence~\citep{bourtoule2021machine}; reversible forgetting retains recoverability and cannot satisfy required erasure. Concept-drift methods detect distributional change~\citep{gama2014survey,weng2024streams}, whereas agent-memory and experience systems govern storage, retrieval, and transfer~\citep{xiong2025memory,tang2025agentkb}.

The closest precedent is the curation loop of \citet{cui2026closing}, which validates and governs verbal insights and highlights stale experience in finance. Knowledge activation packages institutional knowledge into reusable units~\citep{bakal2026activation}, and organizational reviews emphasize governance~\citep{anjelia2025organizational}. HRMC instead makes a specific, falsifiable proposal: active--dormant--retired states, asymmetric thresholds, persistence counters, shadow-tested reactivation, policy-gated retirement, and a transition ledger. We claim these as design hypotheses, not demonstrated advantages.


\section{Why enterprise agents need reversible forgetting}

Enterprise agents accumulate facts, interactions, tools, workflows, and policies~\citep{sapkota2025taxonomy,zheng2025roadmap}. The usual update \(M_{t+1}=M_t\cup\Delta M_t\) mixes memories acquired under incompatible assumptions. This is consequential because organizational knowledge is authoritative, versioned, access-controlled, and tied to owners~\citep{anjelia2025organizational,bakal2026activation}. Yet audits, incident reconstruction, and recurring conditions require historical versions. Reversible forgetting targets this combination of suppression, recoverability, provenance, and governance.


\section{Reversible forgetting framework}

\begin{figure}[t]
\centering
\includegraphics[width=0.80\linewidth]{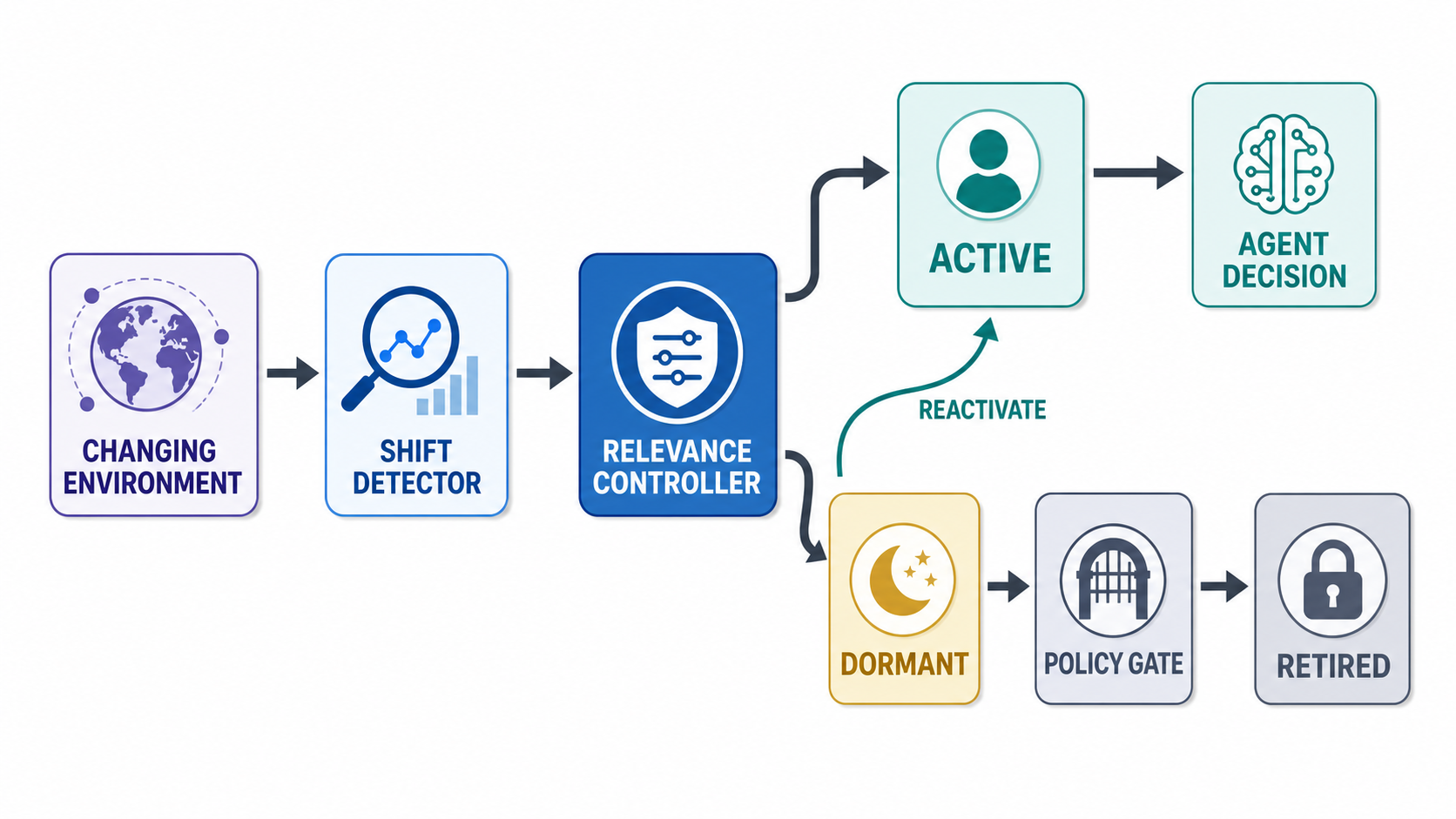}
\caption{Reversible-forgetting lifecycle: change triggers relevance assessment; dormant knowledge can return to active use, whereas retirement requires a policy gate.}
\label{fig:framework}
\end{figure}

We treat enterprise memory as a lifecycle rather than a retention/deletion binary. A shift detector produces context \(R_t\); a controller then assesses which memories should remain active, become dormant, or move toward retirement (Figure~\ref{fig:framework}). Suppression is not deletion.

\paragraph{Memory states.}
\textbf{Active} knowledge supports normal retrieval; \textbf{dormant} knowledge is suppressed but recoverable; \textbf{reactivation} returns it to active use; and \textbf{retired} knowledge cannot be autonomously restored. A provenance stub may remain subject to policy. Retirement is not erasure: legally mandated deletion requires a separate unlearning or records-management process. The objective is to ensure that the \textbf{right knowledge influences the agent at the right time}.

\paragraph{Formalizing relevance.}
For a memory \(m_i\) at time \(t\), consider the conceptual score

\begin{equation}
R_t(m_i)=
\alpha S_t(m_i)+
\beta U_t(m_i)+
\gamma C(m_i)-
\delta A_t(m_i)-
\lambda H_t(m_i),
\label{eq:relevance}
\end{equation}

where \(S\), \(U\), \(C\), \(A\), and \(H\) denote contextual similarity, utility, reliability, staleness, and harm. Implementations might estimate them from embeddings, counterfactual performance, source history, superseding events, and policy violations. The score illustrates candidate signals rather than a validated estimator; proxies and thresholds must be calibrated by memory layer and policy domain.

\subsection{Hysteretic Reversible Memory Controller}

A direct threshold rule can repeatedly move a memory between states. The \textbf{Hysteretic Reversible Memory Controller} (HRMC) instead maintains
\begin{equation}
\bar R_t(m_i)=\rho\bar R_{t-1}(m_i)+(1-\rho)R_t(m_i),
\label{eq:smoothed-relevance}
\end{equation}
with \(\tau_{\downarrow}<\tau_{\uparrow}\). Dormancy requires \(L_{\downarrow}\) consecutive low scores; reactivation requires \(L_{\uparrow}\) high scores and shadow gain above \(\epsilon\). Hysteresis and persistence limit state ``flapping.''

After \(T_{\mathrm{retire}}\), retirement still requires policy or owner approval. Every transition and its evidence enter a ledger. Algorithm~\ref{alg:hrmc} leaves estimators and policies domain-specific.

\begin{algorithm}[H]
\caption{Hysteretic Reversible Memory Controller (HRMC)}
\label{alg:hrmc}
\footnotesize
\begin{algorithmic}[1]
\Require Memory \(m_i\), context \(R_t\), state \(z_i\), score \(\bar R_{t-1}\), counters \(c_i^-\), \(c_i^+\)
\State \(r_i \gets \Call{Relevance}{m_i,R_t}\)
\State \(\bar R_t \gets \rho\bar R_{t-1}+(1-\rho)r_i\)
\If{\(z_i=\textsc{Active}\)}
  \State \(c_i^- \gets \Call{UpdateLowCount}{\bar R_t<\tau_{\downarrow}}\)
  \If{\(c_i^-\ge L_{\downarrow}\) \textbf{or} \(\Call{PolicyBlock}{m_i,R_t}\)}
    \State \(\Call{Transition}{m_i,\textsc{Dormant},evidence}\)
  \EndIf
\ElsIf{\(z_i=\textsc{Dormant}\)}
  \State \(c_i^+ \gets \Call{UpdateHighCount}{\bar R_t>\tau_{\uparrow}}\)
  \If{\(c_i^+\ge L_{\uparrow}\) \textbf{and} \(\Call{ShadowGain}{m_i,R_t}>\epsilon\)}
    \State \(\Call{Transition}{m_i,\textsc{Active},evidence}\)
  \ElsIf{\(\Call{DormantAge}{m_i}>T_{\mathrm{retire}}\) \textbf{and} \(\Call{ApproveRetirement}{m_i}\)}
    \State \(\Call{Transition}{m_i,\textsc{Retired},evidence}\)
  \EndIf
\EndIf
\State \(\Call{AppendLedger}{m_i,z_i,\bar R_t,evidence}\)
\end{algorithmic}
\end{algorithm}

HRMC aims to preserve current-context utility while reducing harmful retrieval, transitions, and storage subject to governance. Similarity may nominate a memory for return, but counterfactual utility must justify reactivation.

\paragraph{Granularity.} Suppression must match the memory layer: retrieval filtering for episodes, versioning for facts and policies, preference decay for tools, retirement for workflows, and parameter adaptation for model behavior. The system must decide both \emph{whether} knowledge is relevant and \emph{how} to reduce its influence.


\section{Finance as an illustrative enterprise domain}

\begin{figure}[H]
\centering
\includegraphics[width=0.72\linewidth]{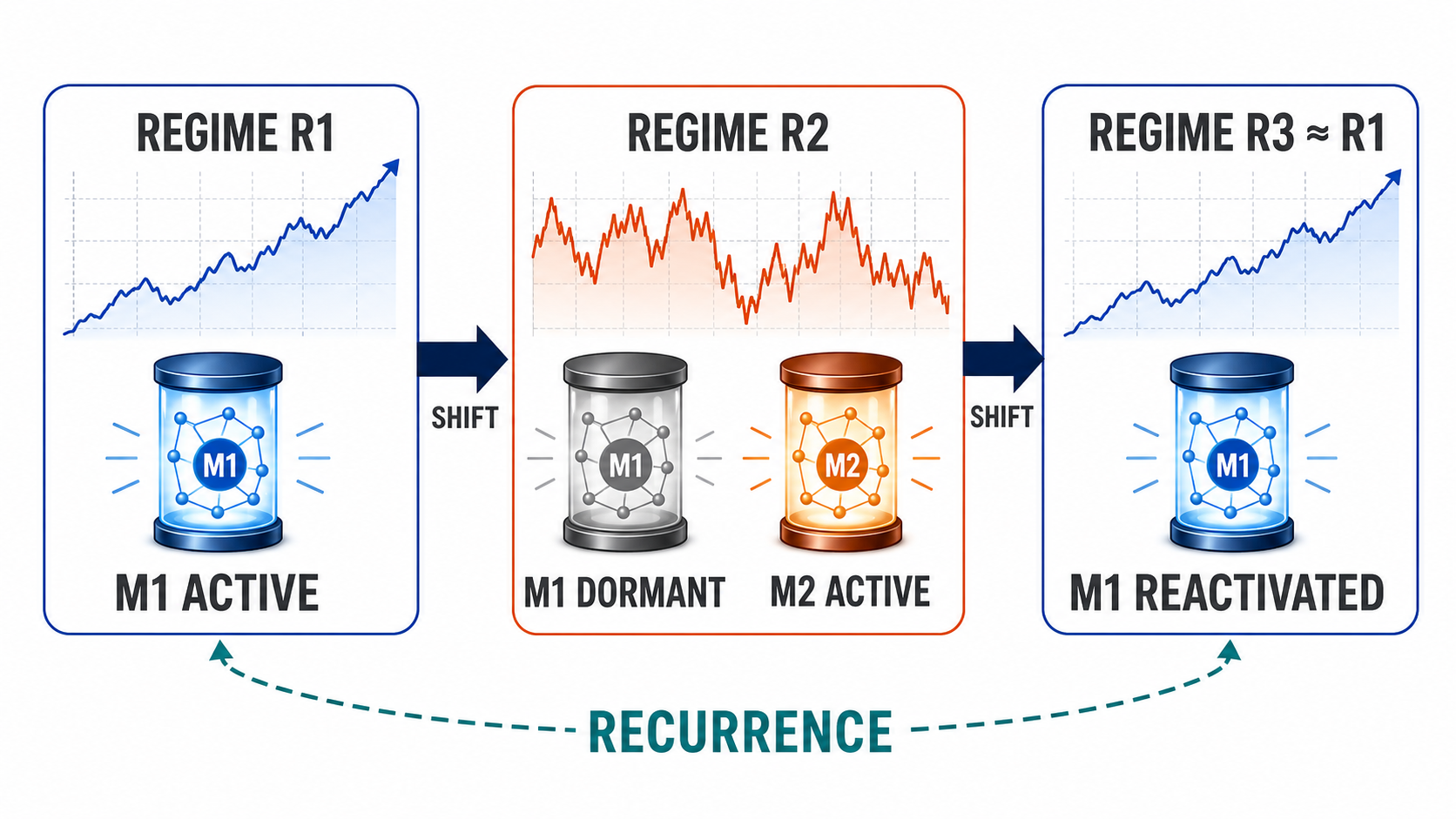}
\caption{Recurring finance regimes: \(M_1\) is active in \(R_1\), becomes dormant while \(M_2\) serves \(R_2\), and is reactivated when \(R_3\) resembles \(R_1\).}
\label{fig:finance}
\end{figure}

Finance makes the idea concrete; stale verbal insights already show costs in this domain~\citep{cui2026closing}. A risk agent may learn thresholds and workflows \(M_1\) in a low-volatility regime \(R_1\). Under crisis regime \(R_2\), these relationships can create excessive alerts. Rather than retain or delete \(M_1\), the controller makes it dormant while \(M_2\) becomes active; if \(R_3\approx R_1\), it reactivates \(M_1\) (Figure~\ref{fig:finance}).

\section{Governance and failure modes}

Enterprise systems must explain why knowledge stopped influencing behavior. A \textbf{forgetting ledger} records

\[
(m_i,\,
state_{\mathrm{old}},\,
state_{\mathrm{new}},\,
reason,\,
time,\,
evidence).
\]

This supports reconstruction, impact analysis, and restoration. Risks include premature suppression, false shift detection, unbounded dormant storage, failed reactivation, adversarial feedback, and compliance conflict. Dormancy cannot satisfy erasure because content remains recoverable. Access control, retention schedules, legal holds, deletion duties, and human override must constrain transitions, aligning the controller with lifecycle risk governance~\citep{nist2023airmf}.


\section{Research and benchmarking agenda}

A practical theory raises six questions: \textbf{detection} of obsolescence; \textbf{granularity} across examples, facts, tools, workflows, policies, and parameters; \textbf{retention} limits for dormant knowledge; \textbf{reactivation} evidence; \textbf{evaluation} separating beneficial suppression from catastrophic forgetting; and \textbf{governance} over transition authority.

Benchmarks need \textbf{recurring environments}. Retention and transfer vary across sequential embodied tasks~\citep{cudrano2024odometry,meng2025robotic}; the sequence \(E_1\!\rightarrow\!E_2\!\rightarrow\!E_3\!\rightarrow\!E_1\) additionally tests whether suppression helps during change and whether prior utility returns upon recurrence.

A benchmark should report \textbf{current-context utility}, \textbf{harmful-retention cost} relative to an oracle policy, \textbf{reactivation recovery}, and \textbf{transition cost}. Governance measures should cover unsupported suppression, policy violations, audit completeness, and actual deletion.

HRMC enables ablations of hysteresis, evidence persistence, shadow evaluation, and policy gating. Retain-all, fixed-window, least-recently-used, permanent-delete, and oracle-context baselines make the position falsifiable: does governed reversible control improve the utility--interference trade-off?

%
\clearpage
\bibliographystyle{plainnat}
\bibliography{references}

@article{kirkpatrick2017overcoming,
  title={Overcoming catastrophic forgetting in neural networks},
  author={Kirkpatrick, James and Pascanu, Razvan and Rabinowitz, Neil and Veness, Joel and Desjardins, Guillaume and Rusu, Andrei A. and Milan, Kieran and Quan, John and Ramalho, Tiago and Grabska-Barwinska, Agnieszka and others},
  journal={Proceedings of the National Academy of Sciences},
  volume={114},
  number={13},
  pages={3521--3526},
  year={2017},
  doi={10.1073/pnas.1611835114}
}

@inproceedings{bourtoule2021machine,
  title={Machine Unlearning},
  author={Bourtoule, Lucas and Chandrasekaran, Varun and Choquette-Choo, Christopher A. and Jia, Hengrui and Travers, Adelin and Zhang, Baiwu and Lie, David and Papernot, Nicolas},
  booktitle={2021 IEEE Symposium on Security and Privacy},
  pages={141--159},
  year={2021},
  doi={10.1109/SP40001.2021.00019}
}

@article{gama2014survey,
  title={A Survey on Concept Drift Adaptation},
  author={Gama, Joao and Zliobaite, Indre and Bifet, Albert and Pechenizkiy, Mykola and Bouchachia, Abdelhamid},
  journal={ACM Computing Surveys},
  volume={46},
  number={4},
  pages={44:1--44:37},
  year={2014},
  doi={10.1145/2523813}
}

@article{xiong2025memory,
  title={How Memory Management Impacts LLM Agents: An Empirical Study of Experience-Following Behavior},
  author={Xiong, Zidi and Lin, Yuping and Xie, Wenya and He, Pengfei and Tang, Jiliang and Lakkaraju, Himabindu and Xiang, Zhen},
  journal={arXiv preprint arXiv:2505.16067},
  year={2025}
}

@techreport{nist2023airmf,
  title={Artificial Intelligence Risk Management Framework ({AI RMF} 1.0)},
  author={Tabassi, Elham},
  institution={National Institute of Standards and Technology},
  number={NIST AI 100-1},
  year={2023},
  doi={10.6028/NIST.AI.100-1}
}

@article{parisi2019continual,
  title={Continual Lifelong Learning with Neural Networks: A Review},
  author={Parisi, German I. and Kemker, Ronald and Part, Jose L. and Kanan, Christopher and Wermter, Stefan},
  journal={Neural Networks},
  volume={113},
  pages={54--71},
  year={2019},
  doi={10.1016/j.neunet.2019.01.012}
}

@article{wang2024comprehensive,
  title={A Comprehensive Survey of Continual Learning: Theory, Method and Application},
  author={Wang, Liyuan and Zhang, Xingxing and Su, Hang and Zhu, Jun},
  journal={IEEE Transactions on Pattern Analysis and Machine Intelligence},
  year={2024},
  doi={10.1109/TPAMI.2024.3367329}
}

@article{delange2022survey,
  title={A Continual Learning Survey: Defying Forgetting in Classification Tasks},
  author={De Lange, Matthias and Aljundi, Rahaf and Masana, Marc and Parisot, Sarah and Jia, Xu and Leonardis, Ales and Slabaugh, Gregory and Tuytelaars, Tinne},
  journal={IEEE Transactions on Pattern Analysis and Machine Intelligence},
  volume={44}, number={7}, pages={3366--3385}, year={2022},
  doi={10.1109/TPAMI.2021.3057446}
}

@article{wickramasinghe2024review,
  title={Continual Learning: A Review of Techniques, Challenges, and Future Directions},
  author={Wickramasinghe, B. and Saha, Gobinda and Roy, Kaushik},
  journal={IEEE Transactions on Artificial Intelligence},
  year={2024}, doi={10.1109/TAI.2023.3339091}
}

@misc{ven2024catastrophic,
  title={Continual Learning and Catastrophic Forgetting},
  author={van de Ven, Gido M. and Soures, Nicholas and Kudithipudi, Dhireesha},
  howpublished={Book chapter}, year={2024},
  doi={10.1016/B978-0-443-15754-7.00073-0}
}

@article{cui2026closing,
  title={Closing the Feedback Loop: From Experience Extraction to Insight Governance in Verbal Reinforcement Learning},
  author={Cui, Ya and Zhang, Xing and Zhang, Yulong and Shao, Lingzhi and Shi, Xiaofeng and Wang, Guanghui and He, Pei-Gen},
  journal={arXiv preprint arXiv:2606.17591}, year={2026},
  doi={10.48550/arXiv.2606.17591}
}

@article{zheng2025roadmap,
  title={Lifelong Learning of Large Language Model Based Agents: A Roadmap},
  author={Zheng, Junhao and Shi, Chengming and Cai, Xidi and Li, Qiuke and Zhang, Duzhen and Li, Chenxing and Yu, Dong and Ma, Qianli},
  journal={IEEE Transactions on Pattern Analysis and Machine Intelligence}, year={2025},
  doi={10.1109/TPAMI.2025.3650546}
}

@phdthesis{weng2024streams,
  title={Handling Non-Stationary Data Streams under Complex Environments},
  author={Weng, Weiwei}, school={Nanyang Technological University}, year={2024},
  doi={10.32657/10356/178601}
}

@article{tang2025agentkb,
  title={{Agent KB}: Leveraging Cross-Domain Experience for Agentic Problem Solving},
  author={Tang, Xiangru and Qin, Tianrui and Peng, Tianhao and Zhou, Ziyang and Shao, Yanjun and Du, TingTing and Wei, Xinming and Xia, Peng and Wu, Fang and Zhu, He and others},
  journal={arXiv preprint arXiv:2507.06229}, year={2025},
  doi={10.48550/arXiv.2507.06229}
}

@article{anjelia2025organizational,
  title={{AI} Agents for Organizational Knowledge Retrieval and Sharing: A Systematic Literature Review},
  author={Anjelia, Sri Rosa and Sensuse, Dana Indra and Lusa, Sofian},
  journal={International Journal of Advances in Data and Information Systems}, year={2025},
  doi={10.59395/ijadis.v6i3.1462}
}

@article{bakal2026activation,
  title={Knowledge Activation: {AI} Skills as the Institutional Knowledge Primitive for Agentic Software Development},
  author={Bakal, G.}, journal={arXiv preprint arXiv:2603.14805}, year={2026},
  doi={10.48550/arXiv.2603.14805}
}

@article{meng2025robotic,
  title={Preserving and Combining Knowledge in Robotic Lifelong Reinforcement Learning},
  author={Meng, Yuan and Bing, Zhenshan and Yao, Xiangtong and Chen, Kejia and Huang, Kai and Gao, Yang and Sun, Fuchun and Knoll, Alois},
  journal={Nature Machine Intelligence}, year={2025},
  doi={10.1038/s42256-025-00983-2}
}

@article{lin2022backward,
  title={Beyond Not-Forgetting: Continual Learning with Backward Knowledge Transfer},
  author={Lin, Sen and Yang, Li and Fan, Deliang and Zhang, Junshan},
  journal={arXiv preprint arXiv:2211.00789}, year={2022},
  doi={10.48550/arXiv.2211.00789}
}

@article{cudrano2024odometry,
  title={The Empirical Impact of Forgetting and Transfer in Continual Visual Odometry},
  author={Cudrano, Paolo and Luo, Xiaoyu and Matteucci, Matteo},
  journal={arXiv preprint arXiv:2406.01797}, year={2024},
  doi={10.48550/arXiv.2406.01797}
}

@article{sapkota2025taxonomy,
  title={{AI} Agents vs. Agentic {AI}: A Conceptual Taxonomy, Applications and Challenges},
  author={Sapkota, Ranjan and Roumeliotis, Konstantinos I. and Karkee, Manoj},
  journal={Information Fusion}, year={2025},
  doi={10.1016/j.inffus.2025.103599}
}

%
%
%

\end{document}